%% file: ms.tex
\pdfoutput=1 

\documentclass[10pt,twocolumn,letterpaper]{article}

\usepackage{wacv}
\usepackage{times}
\usepackage{epsfig}
\usepackage{graphicx}
\usepackage{amsmath}
\usepackage{amssymb}

\usepackage{multirow}
\usepackage{enumitem}%
\usepackage{booktabs}
\usepackage{cite}
\usepackage{float}
\usepackage[font=small]{caption}
\usepackage{subcaption}
\usepackage[table]{xcolor}
\usepackage[T1]{fontenc}

\newif\ifarxiv
\arxivtrue
\ifarxiv
\else
\usepackage{xr}
\fi
\ifarxiv
\else
	\def\wacvPaperID{****} %
\fi

\wacvfinalcopy %

\ifarxiv
	\def\assignedStartPage{1} %
\else
\fi

\ifarxiv
	\usepackage[pagebackref=true,breaklinks=true,colorlinks,bookmarks=false]{hyperref}	
\else
	\ifwacvfinal
		\usepackage[breaklinks=true,colorlinks,bookmarks=false]{hyperref}
	\else
		\usepackage[pagebackref=true,breaklinks=true,colorlinks,bookmarks=false]{hyperref}
	\fi
\fi

\ifarxiv
\else
\fi

\makeatletter
\renewcommand*{\@fnsymbol}[1]{\ensuremath{\ifcase#1\or \dagger\or \dagger\or \ddagger\or
   \mathsection\or \mathparagraph\or \|\or **\or \dagger\dagger
   \or \ddagger\ddagger \else\@ctrerr\fi}}
\makeatother

\begin{document}

\title{Gallery Filter Network for Person Search}
\ifarxiv
    \author{Lucas Jaffe\thanks{University of California, Berkeley. Correspondence to: Lucas Jaffe <lucasjaffe@berkeley.edu>} \and Avideh Zakhor\footnotemark[2]}
\else
    \author{Lucas Jaffe\\
        UC Berkeley\\
        {\tt\small lucasjaffe@berkeley.edu}
        \and
        Avideh Zakhor\\
        UC Berkeley\\
        {\tt\small avz@berkeley.edu}
    }
\fi

\maketitle
\ifarxiv
\else
	\thispagestyle{empty}
\fi

\begin{abstract}
In person search, we aim to localize a query person from one scene in other gallery scenes. The cost of this search operation is dependent on the number of gallery scenes, making it beneficial to reduce the pool of likely scenes. We describe and demonstrate the Gallery Filter Network (GFN), a novel module which can efficiently discard gallery scenes from the search process, and benefit scoring for persons detected in remaining scenes. We show that the GFN is robust under a range of different conditions by testing on different retrieval sets, including cross-camera, occluded, and low-resolution scenarios. In addition, we develop the base SeqNeXt person search model, which improves and simplifies the original SeqNet model. We show that the SeqNeXt+GFN combination yields significant performance gains over other state-of-the-art methods on the standard PRW and CUHK-SYSU person search datasets. To aid experimentation for this and other models, we provide standardized tooling for the data processing and evaluation pipeline typically used for person search research.
\end{abstract}

\section{Introduction}

In the \textit{person search} problem, a \textit{query} person image crop is used to localize co-occurrences in a set of scene images, known as a \textit{gallery}. The problem may be split into two parts: 1) \textit{person detection}, in which all person bounding boxes are localized within each gallery scene and 2) \textit{person re-identification} (re-id), in which detected gallery person crops are compared against a query person crop. Two-step person search methods \cite{zheng_person_2017, chen_person_2018, lan_person_2018, han_re-id_2019, dong_instance_2020, wang_tcts_2020} tackle each of these parts explicitly with separate models. In contrast, end-to-end person search methods \cite{xiao_joint_2017, xiao_ian_2019, liu_neural_2017, ferrari_rcaa_2018, yan_learning_2019, munjal_query-guided_2019, dong_bi-directional_2020, zhong_robust_2020, chen_norm-aware_2020, kim_prototype-guided_2021, li_sequential_2021, han_decoupled_2021, zhang_diverse_2021, yan_anchor-free_2021, yu_cascade_2022, cao_pstr_2022, chen_hierarchical_2020, han_end--end_2021, li_cross-scale_2021} use a single model, typically sharing backbone features for detection and re-identification.

\begin{figure}[t]%
\begin{center}
   \includegraphics[width=1\linewidth]{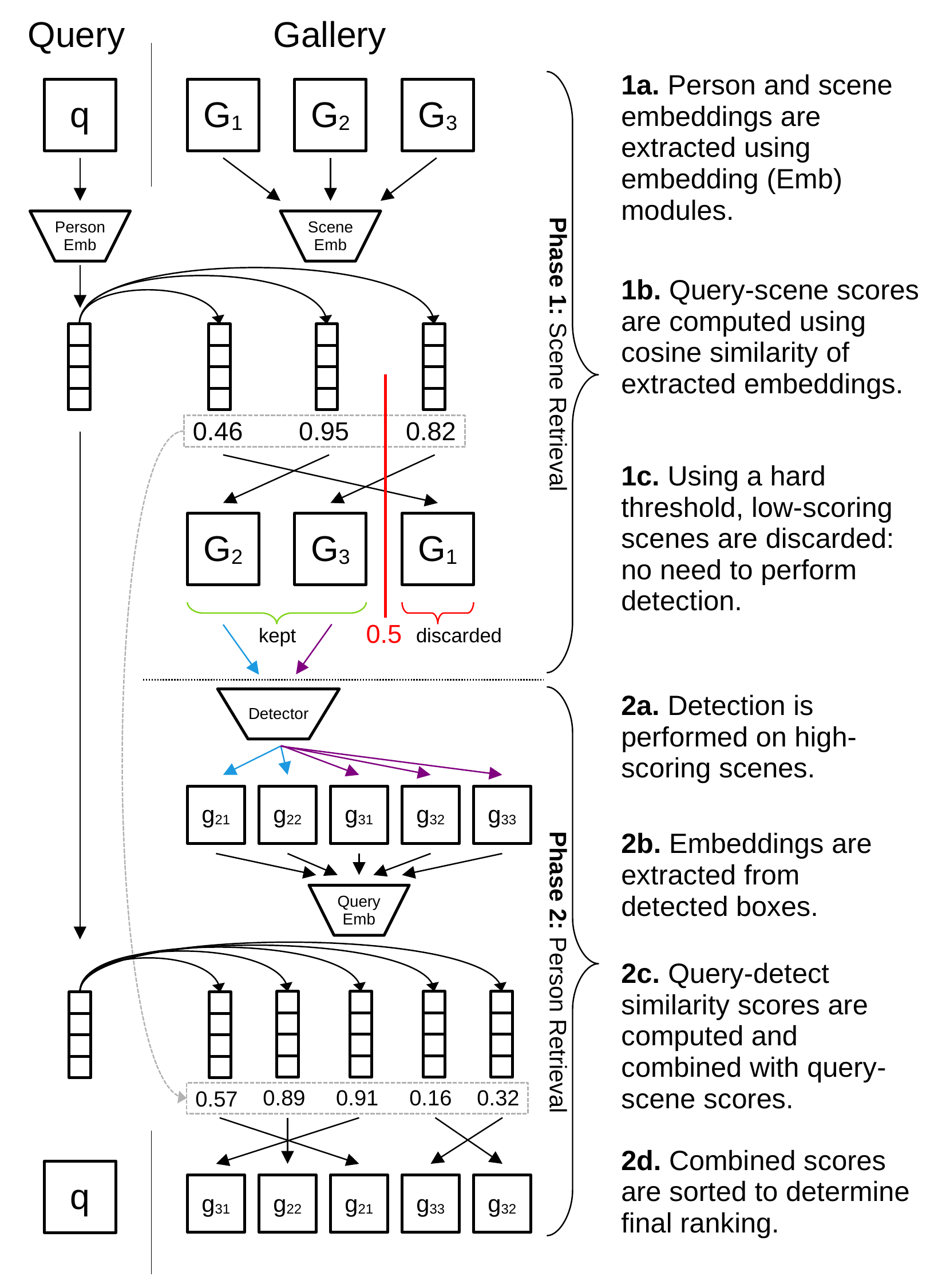}
\end{center}
\vspace{-0.5cm}
\caption{An illustration of our proposed two-phase retrieval inference pipeline. In the first phase, the Gallery Filter Network discards scenes unlikely to contain the query person. The second phase is the standard person retrieval process, in which persons are detected, corresponding embeddings extracted, and these embeddings are compared to the query to produce a ranking.} 
\label{fig:gfn_retrieval}
\end{figure}

For both model types, the same steps are needed: 1) computation of detector backbone features, 2) detection of person bounding boxes, and 3) computation of feature embeddings for each bounding box, to be used for retrieval. Improvement of person search model efficiency is typically focused on reducing the cost of one or more of these steps. We propose the second and third steps can be avoided altogether for some subset of gallery scenes by splitting the retrieval process into two phases: scene retrieval, followed by typical person retrieval. This two-phase process is visualized in Figure \ref{fig:gfn_retrieval}. We call the module implementing scene retrieval the Gallery Filter Network (GFN), since its function is to filter scenes from the gallery.

By performing the cheaper query-scene comparison before detection is needed, the GFN allows for a modular computational pipeline for practical systems, in which one process can determine which scenes are of interest, and another can detect and extract person embeddings only for interesting scenes. This could serve as an efficient filter for video frames in a high frame rate context, or to cheaply reduce the search space when querying large image databases.

The GFN also provides a mechanism to incorporate global context into the gallery ranking process. Instead of combining global context features with intermediate model features as in \cite{dong_instance_2020, li_cross-scale_2021}, we explicitly compare global scene embeddings to query embeddings. The resulting score can be used not only to filter out gallery scenes using a hard threshold, but also to weight predicted box scores for remaining scenes.

We show that both the hard-thresholding and score-weighting mechanisms are effective for the benchmark PRW and CUHK-SYSU datasets, resulting in state-of-the-art retrieval performance (+2.7\% top-1 accuracy on the PRW dataset over previous best model), with improved efficiency (over 50\% per-query cost savings on the CUHK-SYSU dataset vs. same model without the GFN). Additionally, we make contributions to the data processing and evaluation frameworks that are used by most person search methods with publicly available code. That work is described in Supplementary Material Section \ref{supp:data_proc}.

\subsection{Contributions}

Our contributions are as follows:
\begin{itemize}[noitemsep, topsep=0pt]
  \item The Gallery Filter Network: A novel module for learning query-scene similarity scores which efficiently reduces retrieval gallery size via hard-thresholding, while improving detected embedding ranking with global scene information via score-weighting.
  \item Performance improvements and removal of unneeded elements in the SeqNet person search model \cite{li_sequential_2021}, dubbed SeqNeXt.
  \item Standardized tooling for the data pipeline and evaluation frameworks typically used for the PRW and CUHK-SYSU datasets, which is extensible to new datasets.
\end{itemize}

All of our code and model configurations are made publicly available\footnote{Project repository: \url{https://github.com/LukeJaffe/GFN}}.

\section{Related Work}

{\noindent {\bf Person Search.}}
Beginning with the release of two benchmark person search datasets, PRW \cite{zheng_person_2017} and CUHK-SYSU \cite{xiao_joint_2017}, there has been continual development of new deep learning models for person search. Most methods utilize the Online Instance Matching (OIM) Loss from \cite{xiao_joint_2017} for the re-id feature learning objective. Several methods \cite{yan_anchor-free_2021, li_cross-scale_2021, zhang_diverse_2021} enhance this objective using variations of a triplet loss \cite{schroff_facenet_2015}.

Many methods make modifications to the object detection sub-module. In \cite{li_cross-scale_2021, yan_anchor-free_2021, cao_pstr_2022}, a variation of the Feature Pyramid Network (FPN) \cite{lin_feature_2017} is used to produce multi-scale feature maps for detection and re-id. Models in \cite{yan_anchor-free_2021, cao_pstr_2022} are based on the Fully-Convolutional One-Stage (FCOS) detector \cite{tian_fcos_2019}. In COAT \cite{yu_cascade_2022}, a Cascade R-CNN-style \cite{cai_cascade_2018} transformer-augmented \cite{vaswani_attention_2017} detector is used to refine box predictions. We use a variation of the single-scale two-stage Faster R-CNN \cite{ren_faster_2015} approach from the SeqNet model \cite{li_sequential_2021}.

{\noindent {\bf Query-Based Search Space Reduction.}}
In \cite{ferrari_rcaa_2018, liu_neural_2017}, query information is used to iteratively refine the search space within a gallery scene until the query person is localized. In \cite{dong_instance_2020}, Region Proposal Network (RPN) proposals are filtered by similarity to the query, reducing the number of proposals for expensive RoI-Pooled feature computations. Our method uses query features to perform a coarser-grained but more efficient search space reduction by filtering out full scenes before expensive detector features are computed.
{\noindent {\bf Query-Scene Prediction.}}
In the Instance Guided Proposal Network (IGPN) \cite{dong_instance_2020}, a global relation branch is used for binary prediction of query presence in a scene image. This is similar in principal to the GFN prediction, but it is done using expensive intermediate query-scene features, in contrast to our cheaper modular approach to the task.

{\noindent {\bf Backbone Variation.}}
While the original ResNet50 \cite{he_deep_2016} backbone used in SeqNet and most other person search models has been effective to date, many newer architectures have since been introduced. With the recent advent of vision transformers (ViT) \cite{dosovitskiy_image_2021} and a cascade of improvements including the Swin Transformer \cite{liu_swin_2021} and the Pyramid Vision Transformer (v2) \cite{wang_pvt_2022}, used by the PSTR person search model \cite{cao_pstr_2022}, transformer-based feature extraction has increased in popularity. However, there is still an efficiency gap with CNN models, and newer CNNs including ConvNeXt \cite{liu_convnet_2022} have closed the performance gap with ViT-based models, while retaining the inherent efficiency of convolutional layers. For this reason, we explore ConvNeXt for our model backbone as an improvement to ResNet50, which is more efficient than ViT alternatives.

\section{Methods}

\subsection{Base Model}

\begin{figure*}
\begin{center}
\includegraphics[width=1\linewidth]{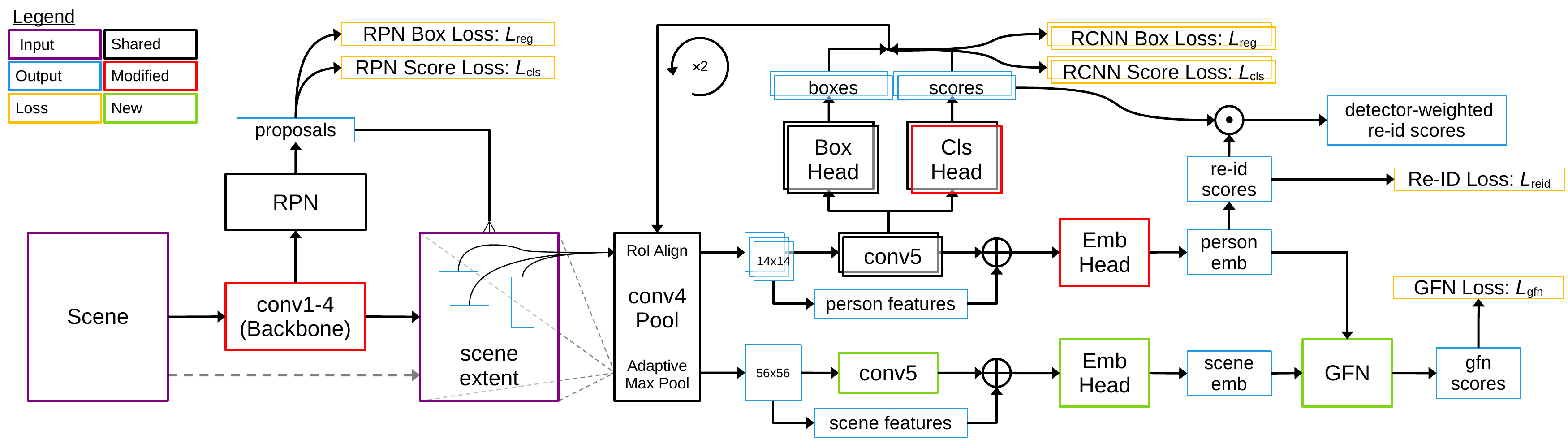}
\end{center}
	\vspace{-0.5cm}
   \caption{Architecture of the SeqNeXt person search model augmented with the GFN. Modules modified from SeqNet are colored red, and new modules, related to the GFN, are colored green. The model follows the standard Faster R-CNN paradigm, with backbone features from \texttt{conv4} being used to generate proposals via the RPN. \texttt{conv4} features are pooled for RPN proposals and passed through the \texttt{conv5} head to generate refined proposals. This process is repeated with the refined proposals to generate the final boxes. \texttt{conv4} features are also used to generate both person embeddings and scene embeddings in the same way: the person box or scene passes through the pooling block and then a duplicated \texttt{conv5} head, and \texttt{conv4}, \texttt{conv5} features are concatenated and passed through an embedding (Emb) head. In the pooling block, RoI Align\cite{he_mask_2020} is used for person and proposal features, while adaptive max pooling is used for scene features. GFN scores are generated using person and scene embeddings from the same or different scenes. Person re-id scores are combined with the score output of the second R-CNN stage to produce detector-weighted scores.}
\label{fig:model}
\end{figure*}

Our base person search model is an end-to-end architecture based on SeqNet \cite{li_sequential_2021}. We make modifications to the model backbone, simplify the two-stage detection pipeline, and improve the training recipe, resulting in superior performance. Since the model inherits heavily from SeqNet, and uses a ConvNeXt base, we refer to it simply as SeqNeXt to distinguish it from the original model. Our model, combined with the GFN module, is shown in Figure \ref{fig:model}.

{\noindent {\bf Backbone Features.}}
Following SeqNet's usage of the first four CNN blocks (\texttt{conv1-4}) from ResNet50 for backbone features, we use the analogous layers in terms of downsampling from ConvNeXt, also referred to as \texttt{conv1-4} for convenience.

{\noindent {\bf Multi-Stage Refinement and Inference.}}
We simplify the detection pipeline of SeqNet by duplicating the Faster R-CNN head \cite{ren_faster_2015} in place of the Norm-Aware Embedding (NAE) head from \cite{chen_norm-aware_2020}. We still weight person similarity scores using the output of the detector, but use the second-stage class score instead of the first-stage as in SeqNet. This is depicted in Figure \ref{fig:model} as ``detector-weighted re-id scores''.

Additionally during inference, we do not use the Context Bipartite Graph Matching (CBGM) algorithm from SeqNet, discussed in Supplementary Material Section \ref{supp:cbgm}.

{\noindent {\bf Augmentation.}}
Following resizing images to 900$\times$1500 (Window Resize) at training time, we employ one of two random cropping methods with equal probability: 1) Random Focused Crop (RFC): randomly take a 512$\times$512 crop in the original image resolution which contains at least one known person, 2) Random Safe Crop (RSC): randomly crop the image such that all persons are contained, then resize to 512$\times$512. This cropping strategy allowed us to train with larger batch sizes, while benefiting performance with improved regularization. At inference time, we resize to 900$\times$1500, as in other models. We also consider a variant of Random Focused Crop (RFC2), which resizes images so the ``focused'' person box is not clipped.

\begin{figure*}
\begin{center}
\includegraphics[width=1\linewidth]{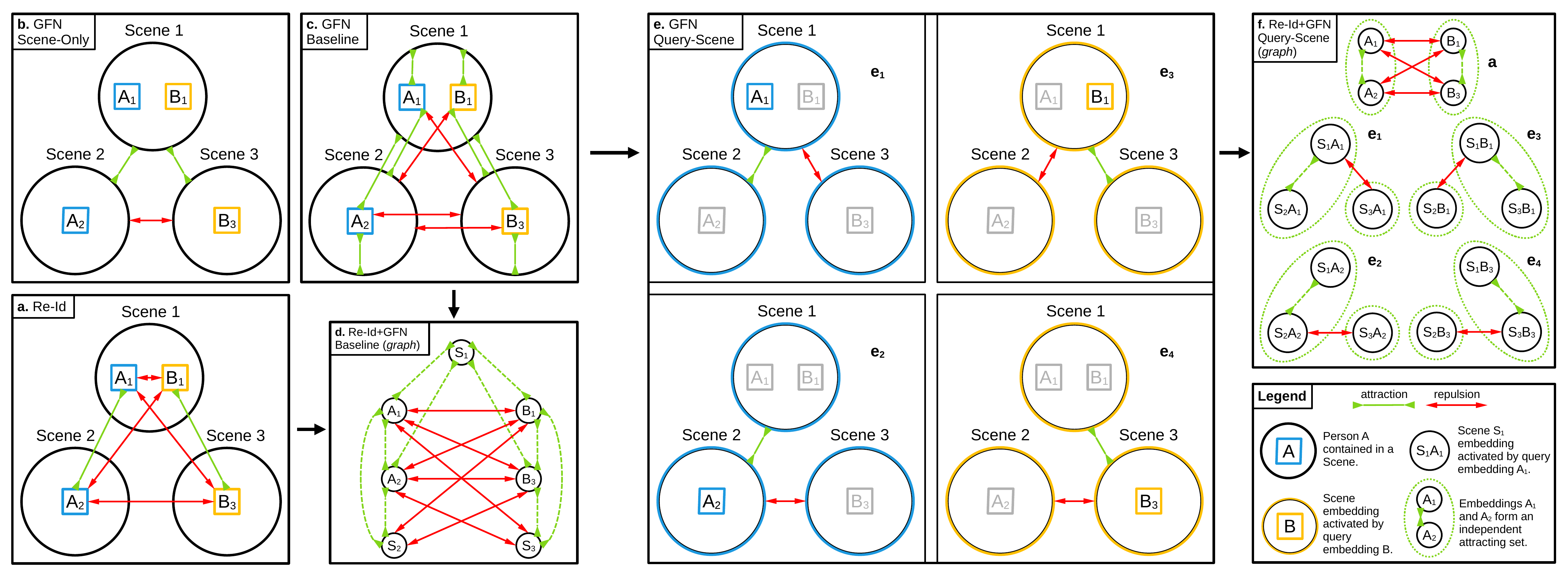}
\end{center}
	\vspace{-0.5cm}
   \caption{Visual representation of the re-id and GFN optimization objectives. In a), b), c), e), circles represent scene images which contain one or more different person identities, labeled A and B. We show a system of three scenes with two unique person identities. Green connectors represent attraction, meaning two embeddings are pushed together by an objective, and red connectors represent repulsion, meaning two embeddings are pulled apart by an objective. In a) we show the standard re-id loss objective. In b) we show the scene-only GFN objective. In c) we show the baseline GFN objective, and in e) we show the combined query-scene GFN objective. In d) we show the graph form of the baseline GFN objective and re-id objective together, and in f) we show the graph form of the combined query-scene GFN objective and re-id objective together, with green ellipses surrounding independent sets in each multipartite component.}
\label{fig:gfn_objective}
\end{figure*}

{\noindent {\bf Objective.}}
As in other person search models, we employ the Online Instance Matching (OIM) Loss \cite{xiao_joint_2017}, represented as $\mathcal{L}_\text{reid}$. This is visualized in Figure \ref{fig:gfn_objective}a. For all diagrams in Figure \ref{fig:gfn_objective}, we borrow from the spring analogy for metric learning used in \textit{DrLIM} \cite{hadsell_dimensionality_2006}, with the concept of \textit{attractions} and \textit{repulsions}.

The detector loss is the sum of classification and box regression losses from the RPN, and the two Faster R-CNN stages, expressed as:

\begin{equation}
\mathcal{L}_\text{det} = \sum_{m \in M} \mathcal{L}^m_\text{cls} + \mathcal{L}^m_\text{reg}, \;\; M = \{\text{\scriptsize RPN}, \text{\scriptsize RCNN1}, \text{\scriptsize RCNN2}\}
\label{eqn:loss_det}
\end{equation}

The full loss is the sum of the detector, re-id, and GFN losses:

\begin{equation}
\mathcal{L} = \mathcal{L}_\text{det} + \mathcal{L}_\text{reid} + \mathcal{L}_\text{gfn}
\label{eqn:loss_full}
\end{equation}

\subsection{Gallery Filter Network}
\label{sec:gfn}
Our goal is to design a module which removes low-scoring scenes, and reweights boxes from higher-scoring scenes. Let $s_\text{reid}$ be the cosine similarity of a predicted gallery box embedding with the query embedding, $s_\text{det}$ be the detector box score, $s_\text{gfn}$ be the cosine similarity for the corresponding gallery scene from the GFN, $\sigma(x) = \frac{e^{-x}}{1+e^{-x}}$, $\alpha$ be a temperature constant, and $\lambda_\text{gfn}$ be the GFN score threshold. At inference time, scenes scoring below $\lambda_\text{gfn}$ are removed, and detection is performed for remaining scenes, with the final score for detected boxes given by \newline $s_\text{final} = s_\text{reid} \cdot s_\text{det} \cdot \sigma(s_\text{gfn} / \alpha)$.

The module should discriminate as many scenes below $\lambda_\text{gfn}$ as possible, while positively impacting the scores of boxes from any remaining scenes. To this end, we consider three variations of the standard contrastive objective \cite{chen_simple_2020, oord_representation_2019} in Sections \ref{sec:gfn_baseline_objective}-\ref{sec:gfn_scene_objective}, in addition to a number of architectural and optimization considerations in Section \ref{sec:gfn_arch_opt}.

\subsubsection{Baseline Objective}
\label{sec:gfn_baseline_objective}

The goal of the baseline GFN optimization is to push person embeddings toward scene embeddings when a person is contained within a scene, and to pull them apart when the person is not in the scene, shown in Figure \ref{fig:gfn_objective}c.

Let $x_i \in \mathbb{R}^d$ denote the embedding extracted from person $q_i$ located in some scene $s_j$. Let $y_j \in \mathbb{R}^d$ denote the embedding extracted from scene $s_j$. Let $X$ be the set of all person embeddings $x_i$, and $Y$ the set of all scene embeddings $y_j$, with $N=|X|, M=|Y|$.

We define the query-scene indicator function to denote positive query-scene pairs as 

\begin{equation}
\mathcal{I}^{Q}_{i,j} = 
    \begin{cases}
        1 & \text{if } q_i \text{ present in } s_j\\
        0 & \text{otherwise}
    \end{cases}
\label{eqn:mem_q}
\end{equation}

We then define a set to denote indices for a specific positive pair and all negative pairs:\newline $K^{Q}_{i,j} = \{ k \in 1,\ldots,M \,|\, k = j \text{ or } \mathcal{I}^{Q}_{i,j} = 0 \}$. Define $\text{sim}(u, v) = u^\top v / \|u\|\|v\|$, the cosine similarity between two $u,v \in \mathbb{R}^d$, and $\tau$ is a temperature constant. 
Then the loss for a positive query-scene pair is the cross-entropy loss

\begin{equation}
\ell^Q_{i,j} = -\log\frac{
	\exp{(\text{sim}(x_i, y_j)/\tau)}
}{
	\sum_{k \in K^{Q}_{i,j}}\exp{(\text{sim}(x_i, y_k)/\tau)}
}
\label{eqn:gfn_p_baseline}
\end{equation}

The baseline Gallery Filter Network loss sums positive pair losses over all query-scene pairs:

\begin{equation}
\mathcal{L}^Q_\text{gfn} = \sum_{i=1}^N \sum_{j=1}^M \mathcal{I}^{Q}_{i,j} \ell^Q_{i,j}
\label{eqn:gfn_l_baseline}
\end{equation}

\subsubsection{Combined Query-Scene Objective}
\label{sec:gfn_qs_objective}
While it is possible to train the GFN directly with person and scene embeddings using the loss in Equation \ref{eqn:gfn_l_baseline}, we show that this objective is ill-posed without modification. The problem is that we have constructed a system of opposing attractions and repulsions. We can formalize this concept by interpreting the system as a graph $G(V, E)$, visualized in Figure \ref{fig:gfn_objective}d. Let the vertices $V$ correspond to person, scene, and/or combined person-scene embeddings, where an edge in $E$ (red arrow) connecting any two nodes in $V$ represents a negative pair used in the optimization objective. Let any group of nodes connected by green dashed arrows (not edges in $G$) be an independent set, representing positive pairs in the optimization objective. Then, each connected component of $G$ must be multipartite, or the optimization problem will be ill-posed by design, as in the baseline objective.

To learn whether a person is contained within a scene while preventing this conflict of attractions and repulsions, we need to apply some unique transformation to query and scene embeddings before the optimization. One such option is to combine a query person embedding separately with the query scene and gallery scene embeddings to produce fused representations. This allows us to disentangle the web of interactions between query and scene embeddings, while still learning the desired relationship, visualized in Figure \ref{fig:gfn_objective}e. The person embedding used to fuse with each scene embedding in a pair is left colored, and the corresponding scenes are colored according to that person embedding. Person embeddings present in scenes which are not used are grayed out.

In the graph-based presentation, shown in Figure \ref{fig:gfn_objective}f, this modified scheme using query-scene embeddings will always result in a graph comprising some number of star graph connected components. Since these star graph components are multipartite by design, the issue of conflicting attractions and repulsions is avoided.

To combine a query and scene embedding into a single query-scene embedding, we define a function $f: \mathbb{R}^d, \mathbb{R}^d \rightarrow \mathbb{R}^d$, such that $z_{i,j} = f(x_i, y_j)$ and $w_i = f(x_i, y^{x_i})$, where $y^{x_i}$ is the embedding of the scene that person $i$ is present in. Borrowing from SENet \cite{hu_squeeze-and-excitation_2018} and QEEPS \cite{munjal_query-guided_2019}, we choose a sigmoid-activated elementwise excitation, with $\odot$ used for elementwise product. ``BN'' is a Batch Normalization layer, to mirror the architecture of the other embedding heads, and $\beta$ is a temperature constant.

\begin{equation}
f(x, y) = \text{BN} ( \sigma(x / \beta) \odot y )
\label{eqn:f_se}
\end{equation}

Other choices are possible for $f$, but the elementwise-product is critical, because it excites the features most relevant to a given query within a scene, eliciting the relationship shown in Figure \ref{fig:gfn_objective}e. %

The loss for a positive query-scene pair is the cross-entropy loss

\begin{equation}
\ell^C_{i,j} = -\log\frac{
	\exp{(\text{sim}(w_i, z_{i,j})/\tau)}
}{
	\sum_{k \in K^{Q}_{i,j}}\exp{(\text{sim}(w_i, z_{i,k})/\tau)}
}
\label{eqn:gfn_p_qs}
\end{equation}

The query-scene combined Gallery Filter Network loss sums positive pair losses over all query-scene pairs:

\begin{equation}
\mathcal{L}^C_\text{gfn} = \sum_{i=1}^N \sum_{j=1}^M \mathcal{I}^{Q}_{i,j} \ell^C_{i,j}
\label{eqn:gfn_l_qs}
\end{equation}

\subsubsection{Scene-Only Objective}
\label{sec:gfn_scene_objective}
As a control for the query-scene objective, we also define a simpler objective which uses scene embeddings only, depicted in Figure \ref{fig:gfn_objective}b. This objective attempts to learn the less discriminative concept of whether two scenes share any persons in common, and has the same optimization issue of conflicting attractions and repulsions as the baseline objective. At inference time, it is used in the same way as the other GFN methods.

We define the scene-scene indicator function to denote positive scene-scene pairs as 

\begin{equation}
\mathcal{I}^{S}_{i,j} = 
    \begin{cases}
        1 & \text{if } s_i \text{ shares any $q$ in common with } s_j\\
        0 & \text{otherwise}
    \end{cases}
\label{eqn:mem_s}
\end{equation}

Similar to Section \ref{sec:gfn_baseline_objective}, we define an index set: \newline $K^{S}_{i,j} = \{ k \in 1,\ldots,M \,|\, k = j \text{ or } \mathcal{I}^{S}_{i,j} = 0 \}$. Then the loss for a positive scene-scene pair is the cross-entropy loss

\begin{equation}
\ell^S_{i,j} = -\log\frac{
	\exp{(\text{sim}(y_i, y_j)/\tau)}
}{
	\sum_{k \in K^{S}_{i,j}}\exp{(\text{sim}(y_i, y_k)/\tau)}
}
\label{eqn:gfn_p_image}
\end{equation}

The scene-only Gallery Filter Network loss sums positive pair losses over all scene-scene pairs:

\begin{equation}
\mathcal{L}^{S}_\text{gfn} = \sum_{i=1}^M \sum_{j=1}^M [i \neq j] \mathcal{I}^{S}_{i,j} \ell^S_{i,j}
\label{eqn:gfn_l_image}
\end{equation}
where $[i \neq j]$ is $1$ if $i \neq j$ else $0$.

\subsubsection{Architecture and Optimization}
\label{sec:gfn_arch_opt}
We consider a number of design choices for the architecture and optimization strategy of the GFN to improve its performance.

{\noindent {\bf Architecture.}}
Scene embeddings are extracted in the same way as person embeddings, except that a larger 56$\times$56 pooling size with adaptive max pooling is used vs. the person pooling size of 14$\times$14 with RoI Align. This larger scene pooling size is needed to adequately summarize scene information, since the scene extent is much larger than a typical person bounding box. In addition, the scene \texttt{conv5} head and Emb Head are duplicated from the corresponding person modules (no weight-sharing), shown in Figure \ref{fig:model}.

{\noindent {\bf Lookup Table.}}
Similar to the methodology used for the OIM objective \cite{xiao_joint_2017}, we use a lookup table (LUT) to store scene and person embeddings from previous batches, refreshing the LUT fully during each epoch. We compare the person and scene embeddings in each batch, which have gradients, with some subset of the embeddings in the LUT, which do not have gradients. Therefore only comparisons of embeddings within the batch, or between the batch and the LUT, have gradients.

{\noindent {\bf Query Prototype Embeddings.}}
Rather than using person embeddings directly from a given batch, we can use the identity prototype embeddings stored in the OIM LUT, similar to \cite{kim_prototype-guided_2021}. To do so, we lookup the corresponding identity for a given batch person identity in the OIM LUT during training, and substitute that into the objective. In doing so, we discard gradients from batch person embeddings, meaning that we only pass gradients through scene embeddings, and therefore only update the scene embedding module. This choice is examined in an ablation in Section \ref{sec:ablation_studies}.

\section{Experiments and Analysis}

\subsection{Datasets and Evaluation}
{\noindent {\bf Datasets.}}
For our experiments, we use the two standard person search datasets, CUHK-SYSU \cite{xiao_joint_2017}, and \textit{Person Re-identification in the Wild} (PRW) \cite{zheng_person_2017}. CUHK-SYSU comprises a mixture of imagery from hand-held cameras, and shots from movies and TV shows, resulting in significant visual diversity. It contains 18,184 scene images annotated with 96,143 person bounding boxes from tracked (known) and untracked (unknown) persons, with 8,432 known identities. PRW comprises video frames from six surveillance cameras at Tsinghua University in Hong Kong. It contains 11,816 scene images annotated with 43,110 person bounding boxes from known and unknown persons, with 932 known identities.

The standard test retrieval partition for the CUHK-SYSU dataset has 2,900 query persons, with a gallery size of 100 scenes per query. The standard test retrieval partition for the PRW dataset has 2,057 query persons, and uses all 6,112 test scenes in the gallery, excluding the identity. For a more robust analysis, we additionally divide the given train set into separate train and validation sets, further discussed in Supplementary Material Section \ref{supp:data_proc}. 

{\noindent {\bf Evaluation Metrics.}} As in other works, we use the standard re-id metrics of mean average precision (mAP), and top-1 accuracy (top-1). For detection metrics, we use recall and average precision at 0.5 IoU (Recall, AP).

In addition, we show GFN metrics mAP and top-1, which are computed as metrics of scene retrieval using GFN scores. To calculate these values, we compute the GFN score for each scene, and consider a gallery scene a match to the query if the query person is present in it.

\subsection{Implementation Details}
We use SGD optimizer with momentum for ResNet models, with starting learning rate 3e-3, and Adam for ConvNeXt models, with starting learning rate 1e-4. We train all models for 30 epochs, reducing the learning rate by a factor of 10 at epochs 15 and 25. Gradients are clipped to norm 10 for all models.

Models are trained on a single Quadro RTX 6000 GPU (24 GB VRAM), and 30 epoch training time using the final model configuration takes 11 hours for the PRW dataset, and 21 hours for the CUHK-SYSU dataset.

Our baseline model used for ablation studies has a ConvNeXt Base backbone, embedding dimension 2,048, scene embedding pool size 56$\times$56, and is trained with 512$\times$512 image crops using the combined cropping strategy (RSC+RFC). It uses the combined prototype feature version of the GFN objective. The final model configuration, used for comparison to other state-of-the-art models, is trained with 640$\times$640 image crops using the altered combined cropping strategy (RSC+RFC2). It uses the combined batch feature version of the GFN objective.

Additional implementation details are given in Supplementary Material Section \ref{supp:add_imp_det}.

\subsection{Comparison to State-of-the-art}
\begin{figure}
\centering
\begin{minipage}[b]{.23\textwidth}
\includegraphics[width=1\linewidth]{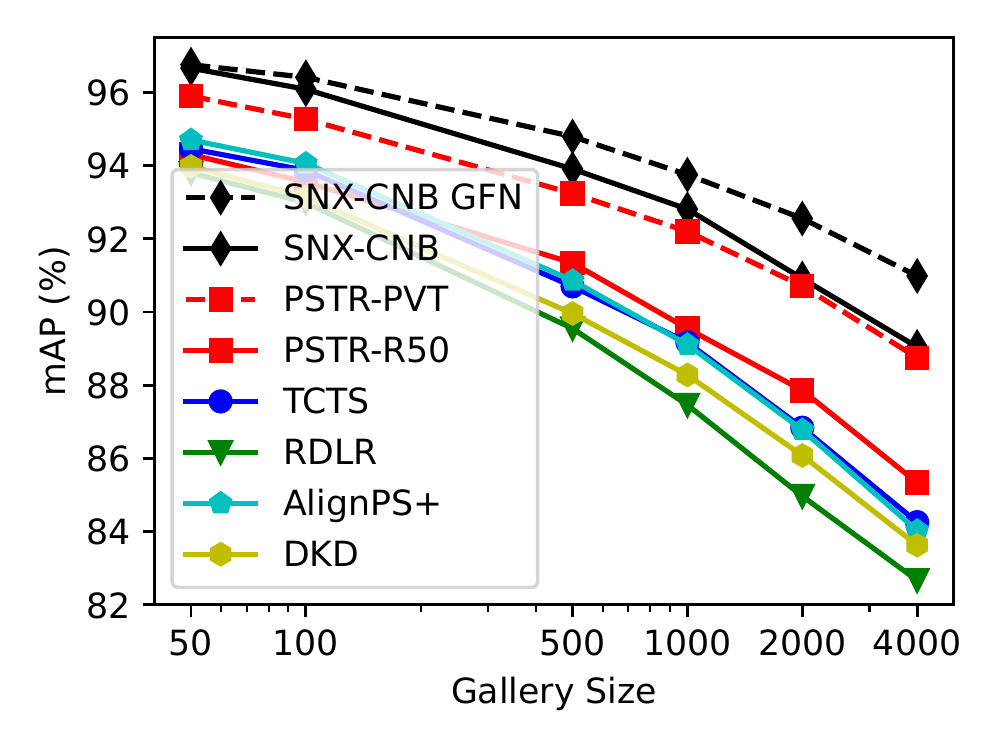}
\vspace{-0.5cm}
\caption{Effect of gallery size on mAP for the CUHK-SYSU dataset. SNX-CNB = SeqNeXt ConvNeXt Base. GFN helps more as gallery size increases.}
\label{fig:cuhk_gallery_size}
\end{minipage}
\hfill
\begin{minipage}[b]{.23\textwidth}
\renewcommand{\arraystretch}{1.25}
\begin{center}
\resizebox{\linewidth}{!}{
\begin{tabular}{|l|cc|}
\hline 
\textbf{Occluded}  & mAP & top-1  \\  
\hline
SeqNeXt &  91.1 &      89.8 \\
SeqNeXt+GFN &         \textbf{92.0} &          \textbf{90.9} \\
\midrule \midrule
\textbf{Low-Resolution}  & mAP & top-1  \\  
\hline
SeqNeXt &     91.4 &      92.4 \\
SeqNeXt+GFN &         \textbf{92.0} &          \textbf{93.1} \\
\hline
\end{tabular}
}
\vspace{-0.2cm}
\end{center}
\captionof{table}{Performance metrics on two CUHK-SYSU retrieval partitions using either Occluded (top) or Low-Resolution (bottom) query persons.}
\label{tab:occlusion_resolution}
\end{minipage}
\end{figure}

We show a comparison of state-of-the-art methods on the standard benchmarks in Table \ref{tab:sota}. The GFN benefits all metrics, especially top-1 accuracy for the PRW dataset, which improves by 4.6\% for the ResNet50 backbone, and 2.9\% for the ConvNeXt Base backbone. Our best model, SeqNeXt+GFN with ConvNext Base, improves mAP by 1.8\% on PRW and 1.2\% on CUHK-SYSU over the previous best PSTR model. This benefit extends to larger gallery sizes for CUHK-SYSU, shown in Figure \ref{fig:cuhk_gallery_size}. In fact, the GFN score-weighting helps more as gallery size increases. This is expected, since the benefit of down-weighting contextually-unlikely scenes, vs. discriminating between persons within a single scene, has a greater effect when there are more scenes compared against.

\begin{table}[t!]
\renewcommand{\arraystretch}{1.0}
\begin{center}
\resizebox{\linewidth}{!}{
\begin{tabular}{|l|c|cc|cc|}
\hline
\multirow{2}*{\textbf{Method}}       & \multirow{2}*{\textbf{Backbone}}      & \multicolumn{2}{c|}{\textbf{CUHK-SYSU}} & \multicolumn{2}{c|}{\textbf{PRW}}  \\\cline{3-6}
       &       & mAP   & top-1 & mAP & top-1  \\
\midrule
\multicolumn{6}{|l|}{\textit{Two-step}}\\
IDE \cite{zheng_person_2017}  & ResNet50    &  - & - & 20.5 & 48.3 \\
MGTS \cite{chen_person_2018}   & VGG16    &  83.0 & 83.7 & 32.6 & 72.1 \\
CLSA \cite{lan_person_2018}  & ResNet50    &  87.2 & 88.5 & 38.7 & 65.0 \\
IGPN  \cite{dong_instance_2020} & ResNet50    &  90.3 & 91.4 & {47.2} & 87.0 \\
RDLR \cite{han_re-id_2019}  & ResNet50    &  93.0 & 94.2 & 42.9 & 70.2 \\
TCTS \cite{wang_tcts_2020}  & ResNet50    &  {93.9} & {95.1} & 46.8 & {87.5} \\
\midrule \midrule
\multicolumn{6}{|l|}{\textit{End-to-end}}  \\
OIM \cite{xiao_joint_2017}      & ResNet50    & 75.5 & 78.7  & 21.3 & 49.4 \\
IAN \cite{xiao_ian_2019}      & ResNet50    & 76.3 & 80.1  & 23.0 & 61.9 \\
NPSM \cite{liu_neural_2017}   & ResNet50    & 77.9 & 81.2  & 24.2 & 53.1 \\
RCAA \cite{ferrari_rcaa_2018}   & ResNet50    & 79.3 & 81.3  & - & - \\
CTXG  \cite{yan_learning_2019}     & ResNet50    & 84.1 & 86.5  & 33.4 & 73.6 \\
QEEPS \cite{munjal_query-guided_2019}      & ResNet50    & 88.9 & 89.1  & 37.1 & 76.7 \\
APNet \cite{zhong_robust_2020}      & ResNet50    & 88.9 & 89.3  & 41.9 & 81.4 \\
HOIM \cite{chen_hierarchical_2020} & ResNet50 & 89.7 & 90.8 & 39.8 & 80.4 \\
BINet \cite{dong_bi-directional_2020}      & ResNet50    & 90.0 & 90.7  & 45.3 & 81.7 \\
NAE+ \cite{chen_norm-aware_2020}       & ResNet50    & 92.1 & 92.9  & 44.0 & 81.1 \\
PGSFL \cite{kim_prototype-guided_2021}   & ResNet50   & 92.3 & 94.7  & 44.2 & 85.2 \\ %
DKD    \cite{zhang_diverse_2021}  & ResNet50    & 93.1 & 94.2  & 50.5 & 87.1 \\
DMRN    \cite{han_decoupled_2021}   & ResNet50    & 93.2 & 94.2  & 46.9 & 83.3 \\
AGWF \cite{han_end--end_2021} & ResNet50 & 93.3 & 94.2 & 53.3 & 87.7 \\
AlignPS \cite{yan_anchor-free_2021}       & ResNet50    & 94.0 & 94.5  & 46.1 & 82.1 \\ %
SeqNet \cite{li_sequential_2021}      & ResNet50    & 93.8 & 94.6  & 46.7 & 83.4 \\
SeqNet+CBGM \cite{li_sequential_2021}      & ResNet50    & 94.8 & 95.7  & 47.6 & 87.6 \\

COAT \cite{yu_cascade_2022} & ResNet50 & 94.2 & 94.7 & 53.3 & 87.4 \\
COAT+CBGM \cite{yu_cascade_2022} & ResNet50 & 94.8 & 95.2 & 54.0 & 89.1 \\
MHGAM \cite{li_cross-scale_2021} & ResNet50 & 94.9 & 95.9 & 47.9 & 88.0 \\
PSTR   \cite{cao_pstr_2022}      & ResNet50    & 94.2 & 95.2 &50.1 &  87.9   \\ %
PSTR  \cite{cao_pstr_2022}      & PVTv2-B2    & 95.2 & 96.2  & 56.5 & 89.7 \\

\hline
SeqNeXt (ours) & ResNet50 & 94.1 &      94.7  & 50.8 &      86.0 \\
SeqNeXt+GFN (ours) & ResNet50 & 94.7 &          95.3 & 51.3 &          90.6 \\
SeqNeXt (ours) & ConvNeXt & 96.1 &      96.5 & 57.6 &      89.5 \\
SeqNeXt+GFN (ours) & ConvNeXt  & \textbf{96.4} &          \textbf{97.0} & \textbf{58.3} &          \textbf{92.4} \\

\hline
\end{tabular}} \vspace{-0.3cm}
\end{center}
\caption{Standard performance metrics mAP and top-1 accuracy on the benchmark CUHK-SYSU and PRW datasets are compared for state-of-the-art \textit{two-step} and \textit{end-to-end} models. ConvNeXt backbone = ConvNeXt Base.} \vspace{-0.4cm}
\label{tab:sota}
\end{table}
\begin{table}[t!]
\renewcommand{\arraystretch}{1.0}
\begin{center}
\resizebox{0.8\linewidth}{!}{
\begin{tabular}{|l|cc|cc|}
\hline
\multirow{2}*{\textbf{Method}} & \multicolumn{2}{c|}{\textbf{Same Cam ID}} & \multicolumn{2}{c|}{\textbf{Cross Cam ID}} \\
\cline{2-5}
 & mAP & top-1  & mAP & top-1  \\
\midrule
HOIM \cite{chen_hierarchical_2020} & - & - & 36.5 & 65.0 \\
NAE+ \cite{chen_norm-aware_2020} & - & -& 40.0 & 67.5 \\
SeqNet \cite{li_sequential_2021} & - & -& 43.6 & 68.5 \\
SeqNet+CBGM \cite{li_sequential_2021} & - & -& 44.3 & 70.6 \\
AGWF \cite{han_end--end_2021} & - & -& 48.0 & 73.2 \\
COAT \cite{yu_cascade_2022} & - & -& 50.9 & 75.1 \\
COAT+CBGM \cite{yu_cascade_2022} & - & -& 51.7 & 76.1 \\
\hline
SeqNeXt (ours) & 82.9 &           98.5 &           55.3 &            80.5 \\
SeqNeXt+GFN (ours) & \textbf{85.1} &           \textbf{98.6} &           \textbf{56.4} &         \textbf{82.1} \\
\hline 
\end{tabular}
}
\vspace{-0.2cm}
\end{center}
\caption{Performance on the PRW test set for query and gallery scenes from the same camera (Same Cam ID) or different cameras (Cross Cam ID).}
\label{tab:cam_id}
\end{table}

The GFN benefits CUHK-SYSU retrieval scenarios with occluded or low-resolution query persons, as shown in Table \ref{tab:occlusion_resolution}. This shows that high quality query person views are not essential to the function of the GFN.

The GFN also benefits both cross-camera and same-camera retrieval, as shown in Table \ref{tab:cam_id}. Strong cross-camera performance shows that the GFN can generalize to varying locations, and does not simply pick the scene which is the most visually similar. Strong same-camera performance shows that the GFN is able to use query information, even when all gallery scenes are contextually similar.

To showcase these benefits, we provide some qualitative results in Supplementary Material Section \ref{supp:qualitative}. These examples show that the GFN uses local person information combined with global context to improve retrieval ranking, even in the presence of difficult confusers.

\subsection{Ablation Studies}
\label{sec:ablation_studies}
We conduct a series of ablations using the PRW dataset to show how detection, re-id, and GFN performance are each impacted by variations in model architecture, data augmentation, and GFN design choices. 

In the corresponding metrics tables, we show re-id results by presenting the GFN-modified scores as mAP and top-1, and the difference between unmodified mAP and top-1 with $\Delta$mAP and $\Delta$top-1. This highlights the change in re-id performance specifically from the GFN score-weighting. To indicate the baseline configuration in a table, we use the $\dagger$ symbol, and the final model configuration is highlighted in gray. 

Results for most of the ablations are shown in Supplementary Material Section \ref{supp:add_ablations}, including model modifications, image augmentation, scene pooling size, embedding dimension, and GFN sampling. 

{\noindent {\bf GFN Objective.}}
We analyze the impact of the various GFN objective choices discussed in Section \ref{sec:gfn}. Comparisons are shown in Table \ref{tab:gfn_objective}. Most importantly, the re-id mAP performance without the GFN is relatively high, but the re-id top-1 performance is much lower than the best GFN methods. Conversely, the Scene-Only method achieves competitive re-id top-1 performance, but reduced re-id mAP.

The Base methods were found to be significantly worse than all other methods, with GFN score-weighting actually reducing GFN performance. The Combined methods were the most effective, better than the Base and Scene-Only methods for both re-id and GFN-only stats, showcasing the improvements discussed in Section \ref{sec:gfn_qs_objective}. In addition, the success of the Combined objective can be explained by two factors: 1) similarity relationship between scene embeddings and 2) query information given by query-scene embeddings. The Scene-Only objective, which uses only similarity between scene embeddings, is functional but not as effective as the Combined objective, which uses both scene similarity and query information. Since the Scene-Only objective incorporates background information, and does not use query information, we reason that the provided additional benefit of the Combined objective comes from the described mechanism of query excitation of scene features, and not from \eg, simple matching of the query background with the gallery scene image.

Finally, the Batch and Proto modifiers to the Combined and Base methods were found to be relatively similar in performance. Since the Proto method is simpler and more efficient, we use it for the baseline model configuration.

\begin{table}[t!]
\renewcommand{\arraystretch}{1.0}
\begin{center}
\resizebox{\linewidth}{!}{
\begin{tabular}{|l|cc|cccc|cc|}
\hline 
  & \multicolumn{2}{|c|}{\textbf{Detection}} & \multicolumn{4}{c|}{\textbf{Re-id}} & \multicolumn{2}{c|}{\textbf{GFN}} \\ 
\midrule
\textbf{GFN Objective} & Recall & AP & mAP & top-1 & $\Delta$ mAP & $\Delta$ top-1 & mAP & top-1 \\
\midrule
None &  96.0 &  \textbf{93.6} &  58.6 &  88.7 & - & - & - & - \\ 
Scene-Only       &           96.0 &           93.4 &           56.5 &           91.9 &           -0.9 &           +2.8 &           16.1 &           73.3 \\
Base Batch  &           95.7 &           93.1 &           53.9 &           86.6 &           -2.6 &           -2.0 &  \textbf{23.8} &           58.4 \\
Base Proto  &           96.0 &           \textbf{93.6} &           55.0 &           86.2 &           -3.0 &           -2.7 &           22.9 &           57.8 \\
\rowcolor{gray!20}
Comb. Batch &  \textbf{96.2} &  \textbf{93.6} &  \textbf{59.5} &           92.2 &  \textbf{+1.1} &           +2.9 &           20.5 &  \textbf{78.8} \\
Comb. Proto$\dagger$ &           96.0 &           93.4 &           58.8 &  \textbf{92.3} &           \textbf{+1.1} &  \textbf{+3.5} &           20.4 &           78.5 \\
\hline 
\end{tabular}
}
\vspace{-0.2cm}
\end{center}
\caption{Comparison of different options for the GFN optimization objective. ``None'' does not use the GFN, Scene-Only uses the objective in Section \ref{sec:gfn_scene_objective}, Base uses the baseline objective in Section \ref{sec:gfn_baseline_objective}, Combined (Comb.) uses the query-scene objective in Section \ref{sec:gfn_qs_objective}, Batch indicates that batch query embeddings are used, Proto indicates that prototype query embeddings are used. Baseline model is marked with \dag, final model is highlighted gray.}
\label{tab:gfn_objective}
\end{table}
\subsection{Filtering Analysis}

\begin{figure}[t]
\begin{center}
   \includegraphics[width=1\linewidth]{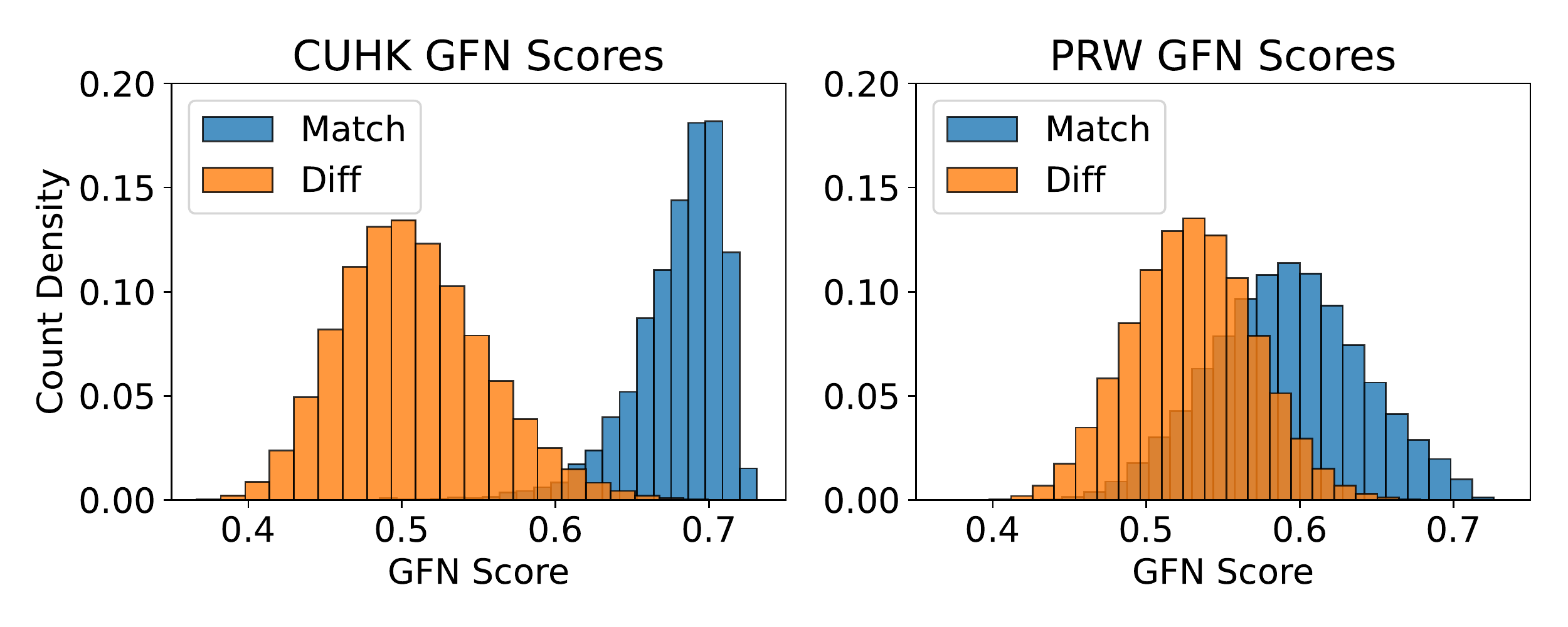}
\end{center}
\vspace{-0.5cm}
\caption{GFN score histograms for the CUHK-SYSU and PRW test sets. Matches and non-matches (Diffs) are shown for queries in the gallery size 4,000 set for CUHK-SYSU, and the full gallery for PRW.}
\label{fig:gfn_histogram}
\end{figure}

{\noindent {\bf GFN Score Threshold.}}
We consider selection of the GFN score threshold value to use for filtering out gallery scenes during retrieval. In Figure \ref{fig:gfn_histogram}, we show histograms of GFN scores for both CUHK-SYSU and PRW. We introduce another metric to help analyze computation savings from the filtering operation: the fraction of negative gallery scenes which can be filtered out (negative predictive value) when using a threshold which keeps 99\% of positive gallery scenes (recall). For the histograms shown, this value is 91.4\% for CUHK-SYSU, and only 11.5\% for PRW.

In short, this is because there is greater variation in scene appearance in CUHK-SYSU than PRW. This results in most query-gallery comparisons for CUHK-SYSU evaluation occurring between scenes from clearly different environments (e.g., two different movies). While the GFN score-weighting improves performance for both same-camera and cross-camera retrieval, shown in Table \ref{tab:cam_id}, query-scene scores used for hard thresholding may be less discriminative for nearly-identical scenes as in PRW vs. CUHK-SYSU, shown in Figure \ref{fig:gfn_histogram}. Still, the GFN top-1 score for the final PRW model was 78.4\%, meaning that 78.4\% of queries resulted in the correct gallery scene being ranked first using only the GFN score.

{\noindent {\bf Compute Cost.}}
In Table \ref{tab:timing}, we show the breakdown of percent time spent on shared computation, GFN-only computation, and detector-only computation. Since most computation time ($\sim$60\%) is spent on detection, with only ($\sim$5\%) of time spent on GFN-related tasks, there is a large cost savings from using the GFN to avoid detection by filtering gallery scenes. Exactly how much time is saved in practice depends on the relative number of queries vs. the gallery size, and how densely populated the gallery scenes are with persons of interest.

To give an understanding of compute savings for a single query, we show some example calculations using the conservative recall requirement of 99\%. For CUHK-SYSU, we have 99.9\% of gallery scenes negative, 91.4\% of negative gallery scenes filtered, and 61.0\% of time spent doing detection on gallery scenes, resulting in 55.7\% computation saved using the GFN compared to the same model without the GFN. For PRW, the same calculation yields 6.6\% computation saved using the GFN.

\begin{table}[t!]
\renewcommand{\arraystretch}{1.2}
\begin{center}
\resizebox{\linewidth}{!}{
\begin{tabular}{|c|c|c|c|c|c|c|}
\hline 
  &\multicolumn{2}{|c|}{\textbf{Shared}} & \multicolumn{2}{c|}{\textbf{GFN}} & \multicolumn{2}{|c|}{\textbf{Detection}} \\
\cline{2-7}
  & Backbone & Query Emb. & Scene Emb. & GFN Scores & RPN & R-CNN($\times$2) \\ 
\hline 
\multirow{2}*{CUHK Time (\%)} & 33.7 & <0.1 & 5.3 & <0.1 & 19.2 & 41.8 \\ 
\cline{2-7}
 & \multicolumn{2}{|c|}{33.7} & \multicolumn{2}{c|}{5.3} & \multicolumn{2}{|c|}{61.0} \\
\midrule \midrule 
\multirow{2}*{PRW Time (\%)} & 36.9 & <0.1 & 5.3 & <0.1 & 16.1 & 41.7 \\ 
\cline{2-7}
 & \multicolumn{2}{|c|}{36.9} & \multicolumn{2}{c|}{5.3} & \multicolumn{2}{|c|}{57.8} \\
\hline

\end{tabular}
}
\vspace{-0.2cm}
\end{center}
\caption{Percent computation time averaged per query of shared feature extraction, GFN, and detection on the CUHK-SYSU (gallery size 4,000) and PRW (gallery size full) test sets.}
\label{tab:timing}
\end{table}

\section{Conclusion}

We describe and demonstrate the Gallery Filter Network, a novel module for improving accuracy and efficiency of person search models. We show that the GFN can efficiently filter gallery scenes under certain conditions, and that it benefits scoring for detects in scenes which are not filtered. We show that the GFN is robust under a range of different conditions by testing on different retrieval sets, including cross-camera, occluded, and low-resolution scenarios. In addition, we show that the benefit given by GFN score-weighting increases as gallery size increases.

Separately, we develop the base SeqNeXt person search model, which has significant performance gains over the original SeqNet model. We offer a corresponding training recipe to train efficiently with improved regularization, using an aggressive cropping strategy. Taken together, the SeqNeXt+GFN combination yields a significant improvement over other state-of-the-art methods. Finally, we note that the GFN is not specific to SeqNeXt, and can be easily combined with other person search models.

{\noindent {\bf Societal Impact.}}
It is important to consider the potential negative impact of person search models, since they are ready-made for surveillance applications. This is highlighted by the PRW dataset being entirely composed of surveillance imagery, and the CUHK-SYSU dataset containing many street-view images of pedestrians.

We consider two potential advantages of advancing person search research, and doing so in an open format. First, that person search models can be used for beneficial applications, including aiding in finding missing persons, and for newly-emerging autonomous systems that interact with humans, \eg, automated vehicles. Second, it allows the research community to understand how the models work at a granular level, and therefore benefits the potential for counteracting negative uses when the technology is abused.

\section*{Acknowledgements}
The authors would like to thank Wesam Sakla and Michael Goldman for helpful discussions and feedback.

\clearpage
{\small
\bibliographystyle{ieee_fullname}
\bibliography{gfn}
}

\ifarxiv
\clearpage
\section*{Supplementary Material}
\input{supplementary.tex}
\fi

\end{document}

%% file: supplementary.tex
\appendix

\section{Data Processing and Evaluation}
\label{supp:data_proc}

We make publicly available our codebase\footnote{\url{https://github.com/LukeJaffe/GFN}}, which includes instructions and config files needed to replicate all main experiments of the paper. For comparitive purposes, we implicitly refer in the following subsections to the public codebases of OIM\footnote{\url{https://github.com/ShuangLI59/person_search}} \cite{xiao_joint_2017}, NAE\footnote{\url{https://github.com/dichen-cd/NAE4PS}} \cite{chen_norm-aware_2020}, SeqNet\footnote{\url{https://github.com/serend1p1ty/SeqNet}} \cite{li_sequential_2021}, COAT\footnote{\url{https://github.com/Kitware/COAT}} \cite{yu_cascade_2022}, AlignPS\footnote{\url{https://github.com/daodaofr/AlignPS}} \cite{yan_anchor-free_2021}, and PSTR\footnote{\url{https://github.com/JialeCao001/PSTR}} \cite{cao_pstr_2022}.

\subsection{Standardized Data Format}
We produce an intermediate COCO-style \cite{lin_microsoft_2014} format for all partitions of the CUHK-SYSU and PRW datatsets. In addition to standard COCO object metadata, we include \texttt{person\_id} and \texttt{is\_known} fields for persons, and a \texttt{cam\_id} image field for performing cross-camera evaluation. 

This standardization process made it straightforward to prepare new partitions of the data. In particular, we split the standard training sets into separate training and validation sets, and created some additional smaller debugging sets. This allowed us to pick hyperparameters without fitting to the test data.

We also standardize the format of retrieval partitions into three categories: 1) fully-specified format which encodes the exact gallery scenes to be used for each query 2) format which specifies queries only, and uses all scenes in the partition as the gallery and 3) format which uses all possible queries, and all possible scenes as the gallery. We create the second and third formats because it is otherwise inefficient to fully-specify the ``all'' cases.

\subsection{Training and Validation Sets}
For both datasets, known identity sets between the train and test partitions are disjoint, making the standard evaluation an \textit{open-set} retrieval problem. To construct the training and validation sets to mirror the open-set retrieval problem of the standard train-test divide, we build a graph based on which scenes share common person identities. Two nodes (scenes) have an edge between them if they share at least one person identity in common. In this way, we can easily split the CUHK-SYSU dataset into a set of connected components, and divide those components into two groups for train ($\sim$80\%) and val ($\sim$20\%).

Since the PRW dataset comprises video surveillance footage, this graph has the property that nearly every scene is connected to another scene via some common person identity. Therefore, we ignore the top 100 most common person identities when constructing the graph for PRW, resulting in a partition which is not quite open-set, but should exhibit similar generalization properties for the purpose of model development. For PRW, we also divide components into two groups for train ($\sim$80\%) and val ($\sim$20\%).

We rename the original train set to ``trainval'', and all of our final experimental results in this paper are from models trained on the full trainval set, and tested on the full test set using the standard retrieval scenarios.

\subsection{Partition Information}
Metadata for the exact breakdown of known and unknown identities and boxes for each partition is given in Table \ref{tab:dataset_info}.

\begin{table}[t!]
\renewcommand{\arraystretch}{1.3}
\begin{center}
\resizebox{\linewidth}{!}{
\begin{tabular}{|l|cccc|cccc|}
\hline 
   & \multicolumn{4}{c|}{\textbf{CUHK-SYSU}} & \multicolumn{4}{c|}{\textbf{PRW}} \\ 
\midrule
\textbf{Metadata}  &  trainval &   test &  train &    val &  trainval &   test &  train &    val \\  
\midrule
Scenes        &       11,206 &    6,978 &     8,964 &   2,242 &            5,704 &        6,112 &         4,563 &       1,141 \\
Boxes         &       55,272 &   40,871 &    44,244 &  11,028 &           18,048 &       25,062 &        14,897 &       3,151 \\
Known IDs     &        5,532 &    2,900 &     4,296 &   1,236 &              483 &          544 &           424 &         158 \\
Known Boxes   &       15,085 &    8,345 &    11,623 &   3,462 &           14,907 &       19,127 &        12,125 &       2,782 \\
Unknown Boxes &       40,187 &   32,526 &    32,621 &   7,566 &            3,141 &        5,935 &         2,772 &         369 \\
\hline 
\end{tabular}
}
\vspace{-0.2cm}
\end{center}
\caption{Dataset metadata showing how many scenes, boxes, and person IDs are in each partition.}
\label{tab:dataset_info}
\end{table}
\subsection{Evaluation Functions}
Using these standardized partitions, we are able to use just one function for detection evaluation and one for retrieval evaluation, as opposed to separate functions for each dataset. This also makes it easier to add in method-specific metrics that can be immediately tested for all partitions.

We note that the current dataset releases for PRW and CUHK-SYSU have a small number (5 or less) of the following errors: duplicate bounding boxes in a single scene, repeated person ids in a single scene, and repeated gallery scenes in a retrieval partition. Although these issues are not handled correctly by the standard evaluation function, we exactly replicate the previous erroneous behavior in our new evaluation function to be certain the comparison against other methods is fair. We leave correction of the underlying data and evaluation function to future work.

\subsection{Augmentation Code Structure}
To make use of augmentation strategies in the albumentations library \cite{buslaev_albumentations_2020}, we refactor evaluation to occur on the augmented data instead of the original data. This allows for easy inclusion of different resizing and cropping strategies which we make use of, in addition to a wealth of other augmentations, experimenting with which we leave to future work.

\subsection{Config Format and Ray Tune}
For running experiments with our code, we provide a YAML config format which is compatible with the Ray Tune library \cite{liaw_tune_2018}. We specifically support the \texttt{tune.grid\_search} functionality by parsing lists in the YAML file as inputs to this function. This makes it easy to run ablations with many variations using a single config file.

\section{Additional Implementation Details}
\label{supp:add_imp_det}
{\noindent {\bf Model Details.}}
We set the OIM scalar (inverse temperature) parameter to 30.0 as in \cite{li_sequential_2021}, with an OIM circular queue size of 5,000 for CUHK-SYSU and 500 for PRW. The OIM momentum parameter is also left at 0.5. For the GFN, the training temperature parameter is 0.1, and the GFN excitation function temperature parameter is 0.2. During training, we use a batch size of 12 for ResNet50 backbone models, and a batch size of 8 for ConvNeXt backbone models.

For the ResNet50 backbone, we freeze all batch norm layers, and all weights through the \texttt{conv1} layer of the model. For ConvNeXt backbones, we freeze only the \texttt{conv1} layer of the model. All backbones are initialized using weights from pre-training on ImageNet1k \cite{russakovsky_imagenet_2015}.

We use automatic-mixed precision (AMP), which significantly reduces all training and inference times. To avoid \texttt{float16} overflow, we refactor all loss functions to divide before summation when computing mean reduction. This increases likelihood of underflow, but results in more stable training overall.

{\noindent {\bf GFN Sampling Strategies.}}
Since we are unable to use the entire GFN LUT to form loss pairs in any given batch due to memory limitations, we have a choice about which LUT embeddings to select for the GFN optimization. By default, for each query person present in the current batch, we sample one matching scene embedding and the person embeddings for all persons in that scene. In addition, we consider sampling a ``hard negative'' scene, defined as a scene which shares at least one person identity in common with the query scene, but that does not contain the query person identity. An ablation for related choices is considered in Section \ref{supp:add_ablations}.

\section{Qualitative Analysis}
\label{supp:qualitative}
\begin{figure*}
\begin{center}
\includegraphics[width=1\linewidth]{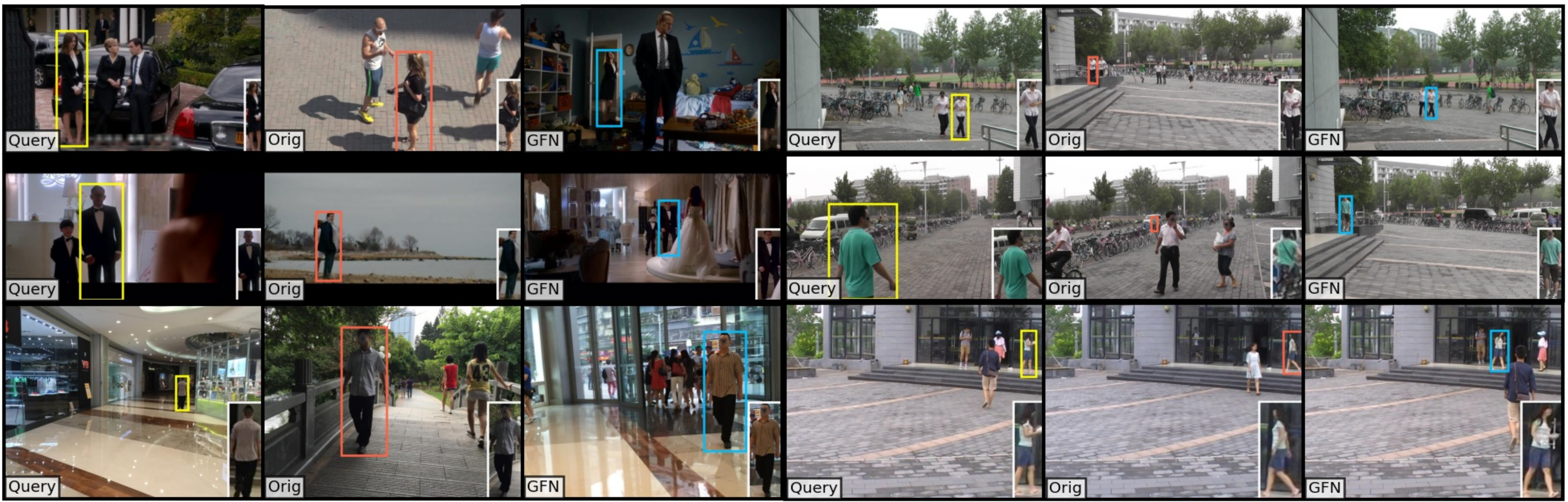}
\end{center}
	\vspace{-0.4cm}
   \caption{Retrieval examples (CUHK-SYSU left, PRW right) from the baseline model where application of the GFN score corrected the top-1 result. The query box is shown in yellow, a false positive gallery match in red, and a true positive gallery match in blue. In each scene, the white box in the lower right duplicates the person of interest for easier comparison between scenes. In the top-left and middle-left, subtle contextual clues (formal wear) help correct the predicted box. In the bottom-left, an obvious contextual clue (interior of same building) corrects the prediction, despite a $180^{\circ}$ change in viewpoint of the person. In the top-right, the false positive and correct match look nearly identical, and the correct box is from the same camera view. In the middle-right, the false positive and correct match have the same shirt and hairstyle, and the correct box is from a different camera view. In the lower-right, the false positive appears to be a mistake in the ground truth (should be true positive), but the GFN ``helped'' by up-weighting a more contextually similar scene.}
\label{fig:qualitative}
\end{figure*}

Qualitative examples are shown for both CUHK-SYSU and PRW in Figure \ref{fig:qualitative}. All examples show cases where the baseline model top-1 match is incorrect, but the GFN-modified match for the same example is correct. We highlight examples where global scene context has an obvious vs. a more subtle impact, and where the query and scene camera ID are the same or different. 

\section{Additional Ablations}
\label{supp:add_ablations}

{\noindent {\bf Model Modifications.}}
We consider how changes to the SeqNet architecture impact performance, including usage of a second Faster R-CNN head instead of the NAE head, and usage of the second detector stage score instead of the first stage score during inference. Results are shown in Table \ref{tab:model_mods}. 

Using the ConvNeXt Base backbone instead of ResNet50 does not improve detection performance, but it significantly improves re-id performance, especially mAP, by 7-8\%. Using the first stage score significantly helps detection performance, but it reduces re-id performance. 

\begin{table}[t!]
\renewcommand{\arraystretch}{1.0}
\begin{center}
\resizebox{\linewidth}{!}{
\begin{tabular}{|l|cc|cccc|cc|}
\hline 
  & \multicolumn{2}{|c|}{\textbf{Detection}} & \multicolumn{4}{c|}{\textbf{Re-id}} & \multicolumn{2}{c|}{\textbf{GFN}} \\ 
\midrule
\textbf{Model} & Recall & AP & mAP & top-1 & $\Delta$ mAP & $\Delta$ top-1 & mAP & top-1 \\  
\midrule
RN50 NAE-FCS  &           97.6 &           93.5 &           50.3 &           89.4 &            0.0 &           +3.5 &           16.7 &           78.5 \\
RN50 RCNN-SCS &           96.0 &           93.1 &           51.1 &           90.6 &           +0.1 &  \textbf{+3.8} &           16.3 &           78.0 \\
CNB NAE-FCS   &  \textbf{97.9} &  \textbf{94.9} &           58.7 &           91.4 &  \textbf{+1.3} &           +3.4 &  \textbf{21.0} &  \textbf{78.9} \\
\rowcolor{gray!20}
CNB RCNN-SCS$\dagger$  &           96.0 &           93.4 &  \textbf{58.8} &  \textbf{92.3} &           +1.1 &           +3.5 &           20.4 &           78.5 \\
\hline 
\end{tabular}
}
\vspace{-0.2cm}
\end{center}
\caption{Comparison of model backbone (RN50=ResNet50, CNB=ConvNeXt Base), NAE vs. R-CNN head for the second detector stage, and first (stage) classifier score (FCS) or second (stage) classifier score (SCS) used at inference time. Baseline model is marked with \dag, final model is highlighted gray.}
\label{tab:model_mods}
\end{table}
{\noindent {\bf Image Augmentation.}}
Shown in Table \ref{tab:augmentation}, we compare the Window Resize augmentation to the two cropping methods used, and a strategy combining the two. We find that the Window Resize method achieves comparable re-id performance with other methods, but much lower detection performance. This may be attributed to the regularizing effect of random cropping for detector training.

In addition, we find that Random Safe Cropping alone results in better detection performance than Random Focused Cropping alone, but worse re-id performance. This shows that the regularizing effect of random crops that may be in the wrong scale is more important for detection, and having features in the target scene scale is more important for re-id. Combining the two results in better performance than either alone for both detection and re-id.

\begin{table}[t!]
\renewcommand{\arraystretch}{1.0}
\begin{center}
\resizebox{\linewidth}{!}{
\begin{tabular}{|l|cc|cccc|cc|}
\hline 
  & \multicolumn{2}{|c|}{\textbf{Detection}} & \multicolumn{4}{c|}{\textbf{Re-id}} & \multicolumn{2}{c|}{\textbf{GFN}} \\ 
\midrule
\textbf{Method} & Recall & AP & mAP & top-1 & $\Delta$ mAP & $\Delta$ top-1 & mAP & top-1 \\  
\midrule
WRS      &           89.3 &           87.7 &           57.3 &           91.1 &           +0.9 &  \textbf{+4.7} &           19.6 &           78.3 \\
RSC      &           95.9 &           93.1 &           55.8 &           91.0 &           +0.7 &           +3.7 &           18.5 &           77.6 \\
RFC      &           95.0 &           92.7 &           58.4 &           91.2 &           \textbf{+1.4} &           +3.4 &           20.8 &           77.8 \\
RFC2     &           95.4 &           93.1 &           58.2 &           91.1 &  \textbf{+1.4} &           +3.8 &  \textbf{21.1} &           78.4 \\
RSC+RFC\dag  &           96.0 &           93.4 &  \textbf{58.8} &  \textbf{92.3} &           +1.1 &           +3.5 &           20.4 &           78.5 \\
\rowcolor{gray!20}
RSC+RFC2 &  \textbf{96.1} &  \textbf{93.8} &           58.7 &  \textbf{92.3} &           +1.3 &           +3.3 &           20.8 &  \textbf{78.9} \\
\midrule
\midrule
\textbf{Crop Size} & Recall & AP & mAP & top-1 & $\Delta$ mAP & $\Delta$ top-1 & mAP & top-1 \\ 
\midrule
256$\times$256 &           95.3 &           91.9 &           51.4 &           90.1 &           0.1 &           3.3 &           16.7 &           78.0 \\
384$\times$384 &  \textbf{96.3} &  \textbf{93.6} &           56.7 &           92.0 &           0.6 &           3.0 &           19.6 &           79.2 \\
512$\times$512\dag &           96.0 &           93.4 &           58.8 &  \textbf{92.3} &           1.1 &  \textbf{3.5} &           20.4 &           78.5 \\
\rowcolor{gray!20}
640$\times$640 &           95.3 &           92.9 &  \textbf{59.6} &           \textbf{92.3} &  \textbf{1.4} &           3.4 &  \textbf{21.8} &  \textbf{79.6} \\
\hline 
\end{tabular}
}
\vspace{-0.2cm}
\end{center}
\caption{Comparison of image augmentation methods (top), and image crop sizes (bottom). Augmentation methods include WRS (Window Resize to 900$\times$1500), RSC (Random Safe Crop to square crop size), RFC (Random Focused Crop to square crop size), RFC2 (variant of RFC), and RSC+RFC(2) which performs either cropping method randomly with equal probability. Baseline model is marked with \dag, final model is highlighted gray.}
\label{tab:augmentation}
\end{table}
{\noindent {\bf Scene Pooling Size and Embedding Dimension.}}
We analyze choices for the RoI Align pooling size for the scene embedding head, and choices for the embedding dimension for both the query and scene embedding heads. Comparisons are shown in Table \ref{tab:feat_pool_size}.

GFN performance increases nearly-monotonically with scene pooling size, with diminishing returns for GFN score-weighted re-id performance. We also note that larger scene pooling size results in a significant increase in memory consumption, so we use 56$\times$56 by default, which captures most of the performance gain, with some memory savings. 

It is clear that the scene pooling size should be larger than the query pooling size to ensure that all person information in a scene is adequately captured. The relationship between person box size distribution vs. scene size, with the ratio of respective pooling sizes could be further investigated. 

For the embedding dimension, performance also increases nearly-monotonically with size, for both re-id and the GFN-only stats. Although there are diminishing returns in performance, like with the scene pooling size, we choose the relatively large value of 2,048 because it results in little additional memory consumption or compute time.

\begin{table}[t!]
\renewcommand{\arraystretch}{1.0}
\begin{center}
\resizebox{\linewidth}{!}{
\begin{tabular}{|l|cc|cccc|cc|}
\hline 
  & \multicolumn{2}{|c|}{\textbf{Detection}} & \multicolumn{4}{c|}{\textbf{Re-id}} & \multicolumn{2}{c|}{\textbf{GFN}} \\ 
\midrule
\textbf{Pool Size} & Recall & AP & mAP & top-1 & $\Delta$ mAP & $\Delta$ top-1 & mAP & top-1 \\ 
\midrule
14$\times$14   &  \textbf{96.1} &           93.5 &           58.1 &           91.6 &           +0.1 &           +3.3 &           18.2 &           77.9 \\
28$\times$28   &           95.9 &           93.4 &           58.5 &           92.3 &           +0.7 &  \textbf{+3.6} &           19.7 &           79.2 \\
\rowcolor{gray!20}
56$\times$56$\dagger$   &           96.0 &           93.4 &           \textbf{58.8} &           92.3 &           +1.1 &           +3.5 &           20.4 &           78.5 \\
112$\times$112 &           \textbf{96.1} &  \textbf{93.6} &  \textbf{58.8} &  \textbf{92.4} &  \textbf{+1.2} &           \textbf{+3.6} &  \textbf{22.1} &  \textbf{79.8} \\
\midrule
\midrule
\textbf{Emb Dim} & Recall & AP & mAP & top-1 & $\Delta$ mAP & $\Delta$ top-1 & mAP & top-1 \\ 
\midrule
128  &           96.1 &           \textbf{93.6} &           58.0 &           91.6 &           0.7 &           3.8 &           19.6 &           77.9 \\
256  &           95.9 &           93.4 &           58.2 &           92.0 &           1.0 &  \textbf{4.3} &           20.1 &           78.3 \\
512  &           96.1 &           93.5 &           58.7 &           91.8 &           1.0 &           4.0 &           20.0 &           77.6 \\
1024 &  \textbf{96.2} &  \textbf{93.6} &  \textbf{59.3} &           92.2 &           \textbf{1.1} &           3.5 &           20.0 &           78.0 \\
\rowcolor{gray!20}
2048$\dagger$ &           96.0 &           93.4 &           58.8 &  \textbf{92.3} &  \textbf{1.1} &           3.5 &  \textbf{20.4} &  \textbf{78.5} \\
\hline 
\end{tabular}
}
\vspace{-0.2cm}
\end{center}
\caption{Comparison of pooling sizes for the RoI Align block used to compute scene embeddings (top) and comparison of the embedding dimension used for both query and scene embeddings (bottom). Baseline model is marked with \dag, final model is highlighted gray.}
\label{tab:feat_pool_size}
\end{table}
{\noindent {\bf GFN Sampling.}}
We analyze choices for the GFN sampling procedure, with comparisons shown in Table \ref{tab:gfn_sampling}. Critically, we find that all sampling options with the LUT are better than not using the LUT at all, as shown by both the large increase in GFN stats, and the contribution of GFN score-weighting to re-id stats. This is expected but important, because it shows that batch-only query-scene comparisons are insufficient (usually just comparing a query to the scene it is present in), and that LUT comparisons are needed despite no gradients flowing through the LUT.

Among sampling mechanisms that use the LUT, results for GFN score-weighted re-id stats were relatively similar, and more trials with more samples per trial are likely needed to distinguish a standout method.

\begin{table}[t!]
\renewcommand{\arraystretch}{1.0}
\begin{center}
\resizebox{\linewidth}{!}{
\begin{tabular}{|l|cc|cccc|cc|}
\hline 
  & \multicolumn{2}{|c|}{\textbf{Detection}} & \multicolumn{4}{c|}{\textbf{Re-id}} & \multicolumn{2}{c|}{\textbf{GFN}} \\ 
\midrule
\textbf{Sampling} & Recall & AP & mAP & top-1 & $\Delta$ mAP & $\Delta$ top-1 & mAP & top-1 \\  
\midrule
No LUT &           \textbf{96.2} &          \textbf{ 93.7} &           57.5 &           90.8 &           -0.3 &           +2.1 &           13.3 &           72.8 \\
P1N0   &           96.1 &           93.6 &  \textbf{59.5} &           91.9 &  \textbf{+1.3} &           +2.4 &           21.0 &           78.7 \\
\rowcolor{gray!20}
P1N1$\dagger$   &           96.0 &           93.4 &           58.8 &  \textbf{92.3} &           +1.1 &           +3.5 &           20.4 &           78.5 \\
P2N0   &  \textbf{96.2} &  \textbf{93.7} &           59.1 &           91.9 &           +1.2 &  \textbf{+3.6} &           20.9 &  \textbf{79.5} \\
P2N1   &           96.0 &           93.6 &           59.1 &           91.7 &           +1.2 &           +3.4 &  \textbf{21.1} &           \textbf{79.5} \\
\hline 
\end{tabular}
}
\vspace{-0.2cm}
\end{center}
\caption{Comparison of different sampling options for optimization of the GFN. P$x$N$y$ indicates that $x$ positive scenes and $y$ hard negative scenes are sampled for each person in the batch. No LUT means we use only batch query and scene embeddings, and no LUT is used. Baseline model is marked with \dag, final model is highlighted gray.}
\label{tab:gfn_sampling}
\end{table}
\section{Comparison with CBGM}
\label{supp:cbgm}
The GFN module is similar to the Context Bipartite Graph Matching (CBGM) method from \cite{li_sequential_2021} in that both methods use context from the query and gallery scenes to improve prediction ranking, although the GFN is used at inference-time only, and does not need to be trained. CBGM is more explicit, in that it directly attempts to match detected person boxes in the query and gallery scenes, at the expense of requiring sensitive hyperparameters: the number of boxes to use from each scene for the matching. The authors found that very different values for these parameters were optimal for the CUHK-SYSU vs. PRW datasets, and did not provide a clear methodology for their selection besides test set performance. In contrast, we use the exact same GFN configuration for both datasets during training and inference, selected separately based on validation data, and found it to robustly improve performance for both.